\documentclass{article}
\usepackage{ijcai25}

\usepackage{times}
\usepackage{soul}
\usepackage{url}
\usepackage[hidelinks]{hyperref}
\usepackage[utf8]{inputenc}
\usepackage[small]{caption}
\usepackage{graphicx}
\usepackage{float}
\usepackage{amsmath}
\usepackage{amsthm}
\usepackage{booktabs}
\usepackage{algorithm}
\usepackage{algorithmic}
\usepackage{multirow} 
\usepackage{amssymb}
\usepackage[switch]{lineno}
\usepackage{bm}
\usepackage[table,xcdraw]{xcolor} 
\usepackage{colortbl} 

\title{Vision and Intention Boost Large Language Model in Long-Term Action Anticipation }

\author{
    Author Name
    \affiliations
    Affiliation
    \emails
    email@example.com
}

\author{
Congqi Cao
\and
Lanshu Hu
\and
Yating Yu
\And
Yanning Zhang\\
\affiliations
Northwestern Polytechnical University
\emails
\{congqi.cao, ynzhang\}@nwpu.edu.cn,
\{hlshu, yatingyu\}@mail.nwpu.edu.cn
}

\begin{document}

\maketitle

\begin{abstract}
    Long-term action anticipation (LTA) aims to predict future actions over an extended period. Previous approaches primarily focus on learning exclusively from video data but lack prior knowledge. Recent researches leverage large language models (LLMs) by utilizing text-based inputs which suffer severe information loss. To tackle these limitations single-modality methods face, we propose a novel Intention-Conditioned Vision-Language (ICVL) model in this study that fully leverages the rich semantic information of visual data and the powerful reasoning capabilities of LLMs. Considering intention as a high-level concept guiding the evolution of actions, we first propose to employ a vision-language model (VLM) to infer behavioral intentions as comprehensive textual features directly from video inputs. The inferred intentions are then fused with visual features through a multi-modality fusion strategy, resulting in intention-enhanced visual representations. These enhanced visual representations, along with textual prompts, are fed into LLM for future action anticipation. Furthermore, we propose an effective example selection strategy jointly considers visual and textual similarities, providing more relevant and informative examples for in-context learning. Extensive experiments with state-of-the-art performance on Ego4D, EPIC-Kitchens-55, and EGTEA GAZE+ datasets fully demonstrate the effectiveness and superiority of the proposed method. 
     
\end{abstract}

\section{Introduction}

\begin{figure}[ht]
	\centering
	\includegraphics[width=0.85\columnwidth]{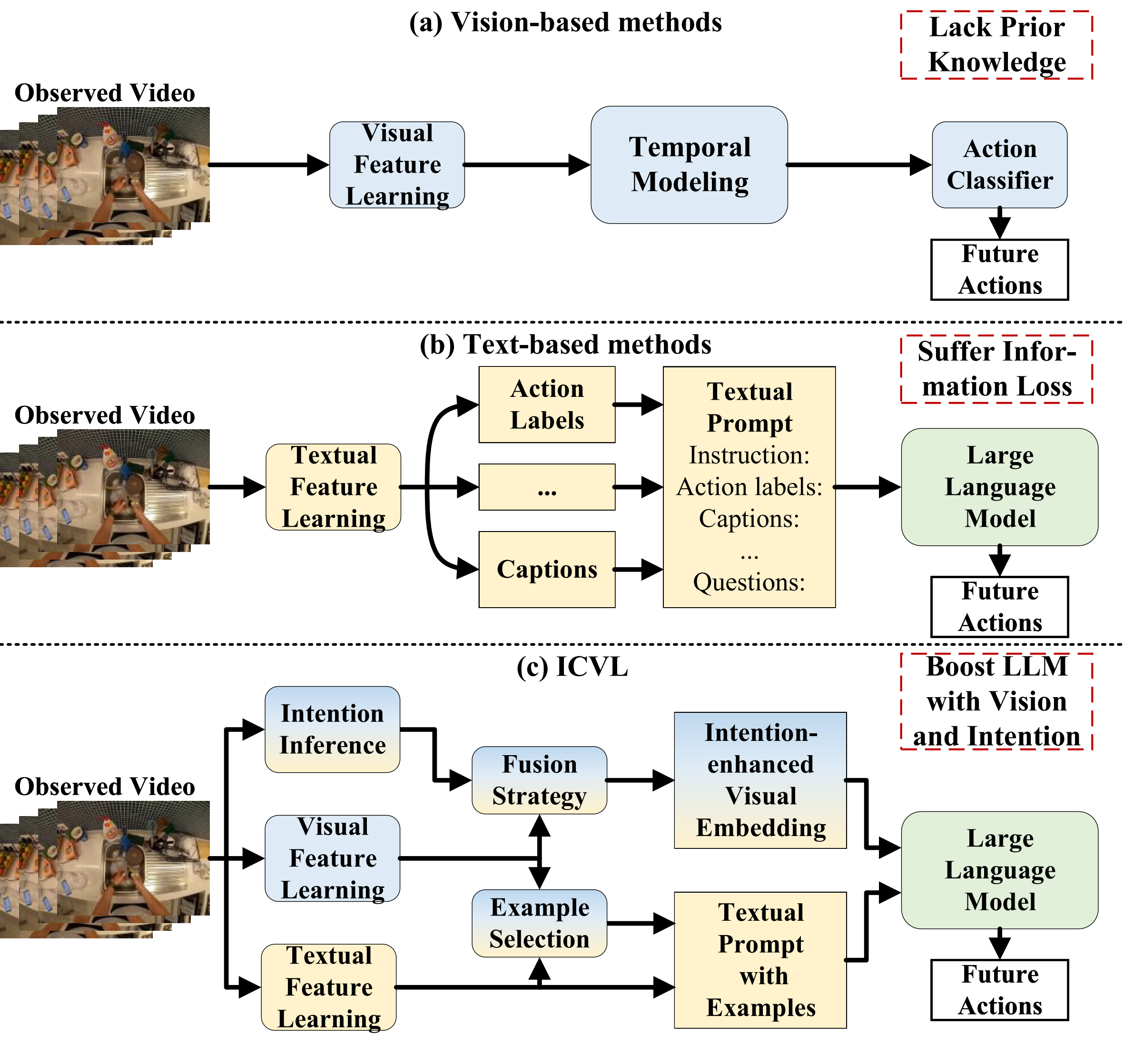} 
	
	\caption{Illustration of different action anticipation methods. (a) Vision-based methods. (b) Text-based methods. (c) Our proposed Intention-Conditioned Vision-Language (ICVL) model.}
	\label{fig:1}
\end{figure}

Predicting future actions is a crucial task in fields such as human-computer interaction and robotic collaboration {\cite{robot:koppula2015anticipating,antirobot:ito2020anticipating}}. This predictive capability enables systems to provide assistance or initiate interactions {\cite{survey:rodin2021predicting,robot:huang2015using}} at the appropriate moments, thereby enhancing both the naturalness and effectiveness of the interaction. For instance, in autonomous driving {\cite{ijcaiad:cao2024sgdcl}}, accurately anticipating the intentions behind the movements of other vehicles enables the autonomous system to make proactive preparations, thereby reducing potential hazards. Unlike other video understanding tasks, action anticipation requires not only understanding the observed context but also predicting future actions based on the observation. This task is inherently challenging, as it requires both strong logical reasoning capabilities and the ability to manage the uncertainty of future actions.

To address the task of action anticipation, some approaches start by leveraging video data to learn visual features and model the temporal relationships between the features via neural networks, as shown in Figure \ref{fig:1} (a). LSTM/RNN-based {\cite{RLSTM:2020rolling,rnn:sener2020temporal,lstm:sadegh2017encouraging}} and Transformer-based {\cite{futr:gong,trans:zhong2023anticipative,mem:wang2023memory}} models are employed to model the dependencies between actions as well as the object-action interaction relationships {\cite{oa:024summarize}}. Furthermore, \cite{cao:2024vs} proposes a hybrid Transformer-GRU architecture to make predictions. However, visual data is often redundant and low in information density. Methods relying solely on visual data lack prior knowledge, making it challenging to model the intrinsic evolution of actions and rendering them overly sensitive to visual variations.

After achieving significant success in natural language processing, large language models (LLMs) {\cite{llm:wang2024visionllm,llmad:cui2024survey,ijcaivlm:yu2023black}} have been adapted to the vision domain, demonstrating remarkable adaptability. This success has motivated researchers to leverage the strong prior knowledge and reasoning capabilities of LLMs to address the challenge of action anticipation. An intuitive solution is to generate appropriate textual substitutes of the original video content, enabling LLMs to predict future actions through a question-answering paradigm, as illustrated in Figure \ref{fig:1} (b). The simplest form of such substitutes is the observed action labels \cite{antgpt:zhao2023antgpt}, generated by off-the-shelf action recognition models. Nevertheless, due to the limited accuracy of existing recognition models, these action labels often contain substantial noise and errors. Another approach involves using a Vision-Language Model (VLM) to generate more detailed textual captions \cite{palm:kim2025palm}. However, fully understanding video content and providing accurate descriptions is inherently challenging. Methods relying solely on textual inputs suffer from significant information loss, limiting the ability of LLMs to make precise and contextually informed predictions.

To fully preserve the visual content and extract crucial clues for long-term action anticipation (LTA), we propose a novel Intention-Conditioned Vision-Language (ICVL) model that integrates complementary visual and textual information with the commonsense prior knowledge of LLMs. This approach addresses the limitations of single-modality methods by combining rich visual features with high-level behavioral intentions to boost the performance of LLMs in LTA. On the one hand, visual data, such as the presence of objects like ``a bowl", offers valuable insights for future action prediction, even when these objects are not explicitly mentioned in textual substitutes. On the other hand, behavioral intentions, such as ``cleaning the kitchen", represent high-level semantic concepts that guide the evolution of actions over time. By capturing these intentions, we can better understand the progression of actions and gain critical insights for predicting future events. 

Specifically, our ICVL model employs a Vision-Language Model (VLM) to infer behavioral intentions directly from video data by analyzing the entire temporal dynamics of the observed video. This allows the model to generate textual features that capture the high-level intentions behind the actions. We then introduce a novel fusion mechanism, Intention-Context Attention Fusion (ICAF), which integrates visual features with the inferred behavioral intentions to produce intention-enhanced visual embeddings. These embeddings are more discriminative, with reduced redundancy and higher information density, as they focus on the most relevant aspects of the visual data guided by behavioral intentions. Combined with carefully designed textual prompts, these enriched embeddings are fed into the LLM, which has been fine-tuned in a parameter-efficient manner to adapt to the specific task of action anticipation. 

Additionally, to further improve the reasoning capabilities of LLMs, we propose an effective example selection mechanism that leverages both visual and textual modalities to identify the most relevant examples for in-context learning. This ensures that the LLM is provided with the most pertinent data, enhancing its ability to make informed and accurate predictions. Extensive experiments across three datasets demonstrate the effectiveness of our approach, validating the strength of combining vision, intention, and LLMs for long-term action anticipation.

Our key contributions can be summarized as follows:
\begin{itemize}
\item We propose a novel multimodal framework for long-term action anticipation that fully leverages both visual and textual information, integrating them with the prior knowledge and reasoning capabilities of LLMs.
\item We introduce intention-enhanced visual features by fusing visual data with inferred behavioral intentions, addressing information loss and enriching the representations for more precise and reliable action predictions.
\item We design an effective example selection mechanism that integrates both visual and textual modalities to identify the most relevant examples, improving long-term action anticipation via enhanced in-context learning.
\item Extensive experiments demonstrate the effectiveness and superiority of our proposed method, achieving state-of-the-art performance on Ego4D, EPIC-Kitchens-55 and EGTEA GAZE+ datasets.
\end{itemize}

\section{Related Works}

\begin{figure*}[t]
	\centering
	\includegraphics[width=\textwidth]{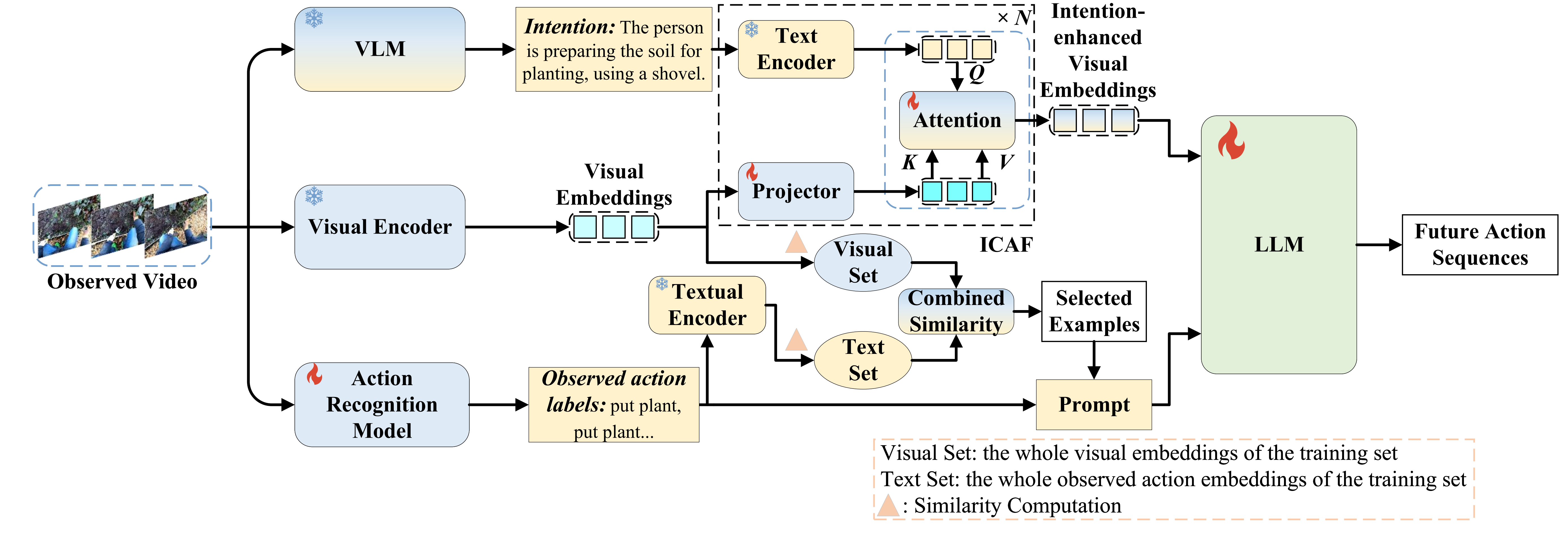} 
	\caption{Illustration of Intention-Conditioned Vision-Language (ICVL) model. Given a video, we use a VLM, a visual encoder, and an action recognition model to extract behavioral intention, original visual embeddings, and observed action labels respectively. The behavioral intention and visual embeddings are then integrated into the intention-enhanced visual embeddings through our proposed Intention-Context Attention Fusion (ICAF) module, in which visual features serve as the keys (\textit{K}) and values (\textit{V}), while textual intention features act as the queries (\textit{Q}). Then we consider both visual similarity and textual similarity based on observed action labels to select examples from the training set for in-context learning. Finally, the textual prompt—composed of instructions, observed action labels, and selected examples—along with the intention-enhanced visual embeddings, are fed into the LLM to generate predictions for future action sequences.}
	\label{fig:2}
\end{figure*}

\subsection{Action Anticipation}
Action anticipation aims at inferring future actions based on a period of observed video, and can be categorized into long-term and short-term anticipation tasks depending on the time to predict. Our work focus on the long-term action anticipation. Previous methods mainly make predictions by modeling the temporal dynamics solely from the visual features. \cite{RLSTM:2020rolling} uses a rolling LSTM to encode the input and an unrolling LSTM to make recurrent predictions for future actions. With the rise of Transformers, \cite{futr:gong} adopts an end-to-end attention model, leveraging fine-grained visual features from previous frames for prediction. Furthermore, \cite{cao:2024vs} proposes a hybrid architecture that utilizes a Transformer as the encoder for long-term sequence modeling, coupled with a GRU decoder for flexible recurrent predictions. Recently, approaches utilizing LLMs have become increasingly popular. \cite{antgpt:zhao2023antgpt} firstly utilizes LLMs to solve the LTA task by simply substituting video content with observed action labels. \cite{palm:kim2025palm} employs an image captioning model to generate descriptions from six aspects of the video, thereby enriching the textual information. Additionally, \cite{egovideo:pei2024egovideo} leverages a more advanced foundation model to extract richer visual features, generating more accurate observed action labels as input for the LLM. However, these methods depend excessively on a single modality. On the one hand, vision-based methods face information redundancy and lack prior knowledge, making it challenging to model long-term temporal relationships and make accurate predictions. On the other hand, text-based methods suffer from severe information loss and noise, struggling to generate accurate action recognition results or detailed video descriptions, thereby impairing predictive accuracy. In contrast to the aforementioned methods, we leverage the contextual information from visual data, the intentional information from textual descriptions, and the commonsense reasoning capabilities of LLMs to enhance long-term action anticipation.

\subsection{Large Language Model}
LLMs based on the Transformer architecture, typically make predictions in an autoregressive manner. These models, often containing billions of parameters, are trained on vast amounts of data and have demonstrated remarkable performance in natural language processing. Notable examples include GPT-4 \cite{gpt4:achiam2023gpt} and LLaMA \cite{llama:touvron2023llama}. To adapt these models more effectively to downstream tasks, some approaches \cite{lora:hu2021lora} propose to fine-tune part of the model parameters on specific datasets, while others \cite{icl:brown2020language} attempt to leverage the LLMs' in-context learning ability by providing high quality examples. Both strategies can further enhance the performance of LLMs, yielding higher-quality responses. Moreover, LLMs exhibit a profound understanding of the textual structure and semantics, as well as the ability to comprehend rich information from other modalities after alignment, such as visual and audio data. As a result, LLMs have been successfully applied to the visual domain, demonstrating significant performance. For example, \cite{llamavid:li2025llama} uses context tokens based on multimodal fusion to represent an entire image and content tokens to encapsulate visual cues for video or image question-answering tasks. Inspired by this approach and the unique nature of LTA task to predict future actions, we creatively leverage high-level behavioral intentions to bridge past and future actions. By combining intentions with visual features, we generate intention-enhanced visual embeddings, which improve prediction by making visual features more discriminative and providing cues related to action evolution.

\section{The Proposed Method: ICVL}

We introduce our proposed Intention-Conditioned Vision-Language (ICVL) model in this section, which combines LLM with intention-enhanced visual embeddings and carefully designed textual prompts to predict future actions, as shown in Figure \ref{fig:2}.

\subsection{Action Recognition and Intention Inference}

\paragraph{Actions labels.} Long-term action anticipation requires predicting future actions over an extended period, where upcoming actions are inferred from an observed video. The observed video can be divided into several segments \text{\{\textit{S}$^i$\}$_{i=1}^{N_{seg}}$}, with each segment \textit{S}$^i$ corresponding to an action label \textit{A}$^i$. Consequently, an observed video can be represented as \text{\{\textit{A}$^1$, \textit{A}$^2$, ..., \textit{A}$^{N_{seg}}$\}}. These action labels are represented as verb-noun pairs, where each action is composed of a verb and a noun \text{\{\textit{v}$^i$, \textit{n}$^i$\}}, such as \textit{put plant}. To make a fair comparison, we follow \cite{antgpt:zhao2023antgpt} and use the CLIP visual encoder to extract video features and get \textit{N}$_{seg}$ visual embeddings represented as \text{\{\textit{E}$^1$, \textit{E}$^2$, ..., \textit{E}$^{N_{seg}}$\}}. Then we use a Transformer-based architecture as the action recognition model, which consists of a Transformer encoder to model the visual embeddings and two MLP heads to decode the verb and noun. For each video segment \textit{S}$^i$, we can obtain the action label based on the identified verb-noun pair. The action recognition model is trained using the cross-entropy loss between predictions and ground-truth action labels. 

\paragraph{Intention Inference.} Human actions are inherently driven by high-level intentions, which guide the evolution of actions over time. Therefore, understanding an individual's intention is crucial for accurately predicting the future actions. While \cite{antgpt:zhao2023antgpt} uses an LLM to infer goals (i.e., intentions) from observed action labels, these labels often contain substantial noise and errors, making it difficult for the LLM to infer correct intentions.  Instead, we leverage observed visual cues through a VLM to obtain more accurate intentions. Specifically, we first uniformly sample \textit{N}$_{frm}$ frames \text{\{\textit{f}$_1$, \textit{f}$_2$, ..., \textit{f}$_{N_{frm}}$\}} from an observed video. These frames can be regarded as a condensed representation of the video’s content, sufficiently indicating the developmental trends of future actions. We then employ a pretrained VLM $\mathcal{E}$ to sequentially infer behavioral intentions \text{\{\textit{I}$_1$, \textit{I}$_2$, ..., \textit{I}$_{N_{frm}}$\}} from each frame in chronological order, using the prompt \textit{P}$^I$ ``\textit{What does the person want to do?}". The intentions inferred from all the preceding frames are also used as input to provide contextual information, supporting the VLM's interpretation of the current image's intention. This can be formulated as:
\begin{equation}
	I_t = \mathcal{E}(P^I, f_t, \sum_{i=1}^{t-1} I_i),
\end{equation}
where $t$ is the index of the current image. The final behavioral intention is derived from the text generated by the VLM based on the last frame and its corresponding context.

\subsection{Intention-Context Attention Fusion}
Multi-modality fusion has been proven effective in short-term action anticipation tasks \cite{RLSTM:2020rolling,cao:2024vs}. However, in the field of long-term action anticipation, this approach remains underexplored, particularly for LLM-based methods. In this section, we introduce our proposed Intention-Context Attention Fusion strategy. 

\paragraph{Visual and Intention embeddings.} For each video segment \textit{S}$^i$, we use a pretrained vision encoder to extract the original visual embeddings through \textit{k} uniformly sampled video frames, resulting in $E^i_{v} \in \mathbf{R}^{k\times d_v}$. The $N_{seg}$ video segments' visual embeddings can be concatenated to represent the whole video's embeddings $E_{v}^{'} \in \mathbf{R}^{T\times d_v}$ where $T = N\times k$. These visual embeddings serve as visual prompts, which are integrated with textual intention prompts as input for the LLM. To enhance the model’s understanding of sequential information, we add 2D fixed positional encoding \cite{test:vaswani2017attention} to the visual embeddings. Then we can get the modified visual embeddings $E_{v}^{''}$ after adding the positional embeddings. Furthermore, to align the dimension of visual embeddings with the embedding space $d_l$ of the LLM, a linear project layer is used to get the final visual embeddings $E_{v} \in \mathbf{R}^{T\times d_l}$. 

For the intention embeddings, a pretrained text encoder can be directly used to encode the behavioral intentions, denoted as $E_{i} \in \mathbf{R}^{seq\times d_l}$, where \textit{seq} and \textit{d}$_l$ represent the sequence length of the intention embeddings and the embedding dimension of the LLM, respectively. 

\paragraph{Fusion strategy.} Our fusion strategy, based on cross-attention, integrates both visual and intention embeddings to obtain enhanced intention-enhanced visual embeddings $E_{ic} \in \mathbf{R}^{seq\times d_l}$. This process can be formulated as:

\begin{equation}
	E_{ic} = \text{Attention}(E_{i}, E_{v}, E_{v}),
\end{equation}

\begin{equation}
	\text{Attention}(Q, K, V) = \text{softmax}\left(\frac{QK^{T}}{\sqrt{d_{l}}}\right)V,
\end{equation}
where visual embeddings serve as keys (\textit{K}) and values (\textit{V}), intention embeddings act as queries (\textit{Q}). And scaling factor $\sqrt{d_{l}}$ is introduced to avoid gradient vanishing. As intention is embedded in visual features and guides the evolution of actions, it can enhance the visual embeddings to be more discriminative. Besides, this fusion also enhances the LLM's ability to comprehend visual information and improves its interpretability, resulting in more discriminative and intention-consistent cross-modal representations.

\subsection{Example Selection}
Existing research \cite{icl:brown2020language} has shown that augmenting LLMs with relative demonstration examples can significantly enhance their generative capabilities. However, selecting appropriate examples for action anticipation tasks remains challenging due to the diversity of scenarios and the variability of actions even within the same scenario. To address this, we propose an example selection mechanism that jointly considers both visual and textual modalities as shown in Figure \ref{fig:2}. And this mechanism can provide more relevant and appropriate examples for in-context learning, thereby improving generalizability. We first introduce single-modality selection and then explain how to extend it to a comprehensive multi-modality approach.

\paragraph{Single-Modality Selection.}
Taking the visual modality selection as an example, after obtaining the whole original visual embeddings $E_{v}^{'}$, we apply average pooling to derive the averaged visual embeddings $\bar{E_v}$ as a global representation of visual features. We then utilize L2 distance to obtain the similarity scores $s_v$ between the query video and all the training videos, due to its clear geometric interpretation and effectiveness in capturing variations in both vector magnitude and feature dimensions, where a smaller similarity score $s_{vi}$ between two features indicates greater similarity. Finally, we select the top-k examples based on the similarity scores. This process can be formulated as below:

\begin{equation}
	U = \arg\min_{U \subset \Omega, |U|=k} \sum_{\bar{E_v^i} \in \Omega} \| E_q - \bar{E_v^i} \|_2,
\end{equation}
where $\Omega$ represents the complete set of $\bar{E_v}$ in the training set, \textit{E}$_{q}$ represents the embeddings of the query and \textit{U} represents the set of the top-k most similar selected embeddings. Based on \textit{U}, corresponding examples are then extracted from the training set, comprising observed action labels and future action sequences. 

The example selection mechanism for the textual modality adheres to the same principles as that of the visual modality, where observed action labels are encoded to obtain textual embeddings.

\begin{figure}[t]
	\centering
	\includegraphics[width=1\columnwidth]{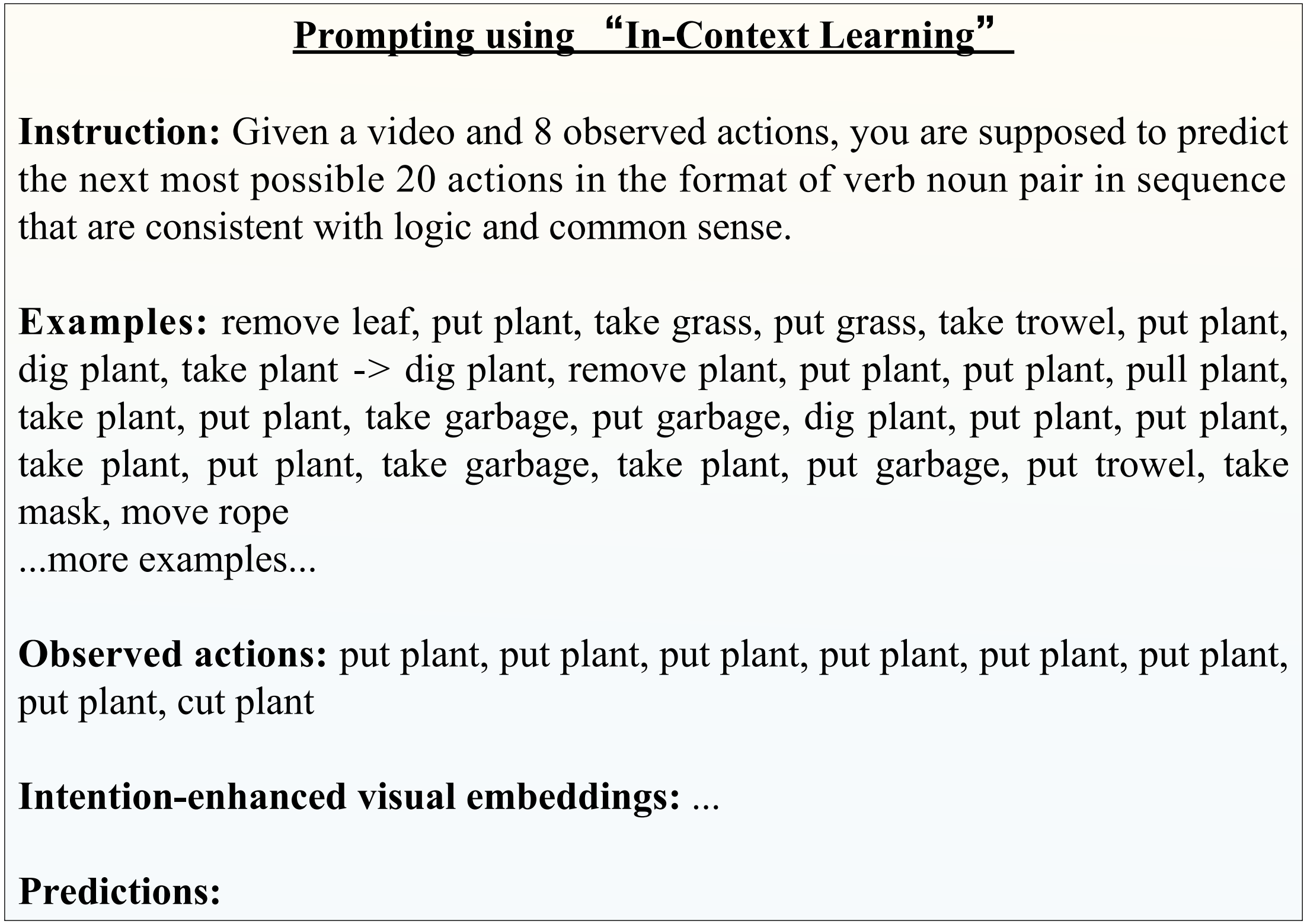} 
	\caption{Illustration of prompt for LLMs using in-context learning. The prompt is composed of an instruction, selected examples based on multi-modality similarity, observed actions and intention-enhanced visual embeddings.}
	\label{fig:prompt}
\end{figure}

\paragraph{Multi-Modality Selection.}
After obtaining the similarity results of the visual and textual modalities, we adopt a weighted summation approach to comprehensively consider the similarities of both modalities. First, the similarity scores for the visual and textual modalities are normalized as shown in the following formula:
\begin{equation}
	s_{ti}^n = \frac{s_{ti} - \min(s_t)}{\max(s_t) - \min(s_t)},
\end{equation}

\begin{equation}
	s_{vi}^n = \frac{s_{vi} - \min(s_v)}{\max(s_v) - \min(s_v)},
\end{equation}
where $s_{ti}^n$ represents the normalized similarity score for the textual modality of the $i$-th representation, and $s_{vi}^n$ represents the normalized similarity score for the visual modality of the $i$-th representation. 

The comprehensive similarity score is then calculated using a weighted summation based on $s_{ti}^n$ and $s_{vi}^n$:

\begin{equation}
	s_i = \alpha \times s_{ti}^n + (1 - \alpha) \times s_{vi}^n,
\end{equation}
where $\alpha$ is a weighting factor that reflects the balance between the two modalities. When $\alpha$ is set to either 1 or 0, the selection mechanism becomes solely dependent on a single modality. Based on the comprehensive similarity scores, the top-k examples are selected. The final prompt is illustrated in the Figure \ref{fig:prompt}.

\begin{table*}[t!]
	\centering
	\begin{tabular}{lccccc}
		\hline
		Method & Venue & Visual Encoder & Noun $\downarrow$ & Verb $\downarrow$ & Action $\downarrow$ \\
		\hline
		PaMsEgoAI \cite{pams:ishibashi2023technical}& arXiv'23 & - & 0.6291 & 0.6702 & 0.8753 \\
		HAI-PUI \cite{mamba:zhong2024querymamba} & arXiv'24 & - & 0.6733 & 0.7721 & 0.9242 \\ 
		\rowcolor{gray!10} 
		AntGPT \cite{antgpt:zhao2023antgpt} & ICLR'23 & CLIP & 0.6755 & 0.6728 & 0.8931 \\
		\rowcolor{gray!10} 
		PlausiVL* \cite{plausivl:mittal2024can} & CVPR'24 & - & 0.6466 & 0.6618 & 0.8771 \\ 
		\rowcolor{gray!10} 
		EgoVideo \cite{egovideo:pei2024egovideo} & arXiv'24 & EgoVideo-V  & \underline{0.6264} & \underline{0.6576} & \underline{0.8619} \\
		\rowcolor{gray!10} 
		PALM \cite{palm:kim2025palm} & ECCV'24 & EgoVLP & 0.6465 & 0.7111 & 0.8819 \\
		\hline
		\rowcolor{gray!10} 
		ICVL(Ours) & - & CLIP & \textbf{0.6194} & \textbf{0.6516} & \textbf{0.8570} \\
		\hline
	\end{tabular}
	\caption{Long-term action anticipation performance on Ego4D. The results with \textbf{bold} and \underline{underline} indicate the highest and second-highest values, respectively. * denotes our reproduced results. Rows with gray shading represent LLM-based method. Visual Encoder refers to the visual encoder of the action recognition model.}
	\label{tab:ego4d}
\end{table*}

\begin{table*}[t!]
	\centering
	\begin{tabular}{lccccccc}
		\toprule
		\multirow{2}{*}{Method} & \multirow{2}{*}{Venue} & \multicolumn{3}{c}{EK-55} & \multicolumn{3}{c}{EGTEA} \\ 
		\cmidrule(lr){3-5} \cmidrule(lr){6-8} 
		&  & ALL $\uparrow$ & FREQ $\uparrow$ & RARE $\uparrow$ & ALL $\uparrow$ & FREQ $\uparrow$ & RARE $\uparrow$ \\ 
		\midrule
		Timeception \cite{timeception:hussein2019timeception} & CVPR'19 & 35.6 & 55.9 & 26.1 & 74.1 & 79.7 & 59.7 \\
		VideoGraph \cite{videograph:hussein2019videograph} & arXiv'19 & 22.5 & 49.4 & 14.0 & 67.7 & 77.1 & 47.2 \\
		EGO-TOPO \cite{egotopo:nagarajan2020ego} & CVPR'20 & 38.0 & 56.9 & 29.2 & 73.5 & 80.7 & 54.7 \\
		Anticipatr \cite{anticipatr:nawhal2022rethinking} & ECCV'22 & 39.1 & 58.1 & 29.1 & 76.8 & 83.3 & 55.1 \\
		\rowcolor{gray!10} 
		AntGPT \cite{antgpt:zhao2023antgpt} & ICLR'23 & 40.1 & 58.8 & \underline{31.9} & 80.2 & 84.8 & 72.9 \\
		\rowcolor{gray!10} 
		PALM \cite{palm:kim2025palm} & ECCV'24 & \underline{40.4} & \underline{59.3} & 30.3 & \underline{80.7} & \underline{85.0} & \underline{73.5} \\
		\hline
		\rowcolor{gray!10} 
		ICVL (Ours) & - & \textbf{43.3} & \textbf{61.6} & \textbf{33.8} & \textbf{81.0} & \textbf{85.2} & \textbf{73.7} \\
		\bottomrule
	\end{tabular}
	\caption{Long-term action anticipation performance on EK-55 and EGTEA datasets. The results with \textbf{bold} and \underline{underline} indicate the highest and second-highest values, respectively. Rows with gray shading represent LLM-based method.}
	\label{tab:ek}
\end{table*}

\subsection{Training}
As shown in Figure \ref{fig:2}, the visual and textual encoders in ICVL are frozen, while the ICAF module are fully trainable. Given the significant computational cost of fully training LLMs, we adopt the LoRA (Low-Rank Adaptation) \cite{lora:hu2021lora} for fine-tuning the LLM. All trainable parameters are optimized based on the text generated by the LLM. As the model is tasked with predicting a future action sequence, We employ the next-token prediction loss with negative log-likelihood to optimize the predicted tokens:

\begin{equation}
	\mathcal{L}_{CE}(\theta) = -\sum_{t=1}^{M} \log p_{\theta}(y_t \mid y_{<t}),
\end{equation}
where $\theta$ represents the parameters of the model, $y_t$ denotes the target token to be predicted at position $t$, $M$ refers to the total number of tokens to be predicted, $y_{<t}$ represents the tokens predicted prior to position $t$, and  $p_\theta$ indicates the probability of successfully predicting the token at position $t$ based on $y_{<t}$. This loss measures the difference between the action sequence output by LLM and the corresponding ground truth action sequence.

We adopt an end-to-end training process that both the ICAF module and the LoRA Adapter module are fine-tuned simultaneously.

\section{Experiment}
In this section, we first introduce the datasets and evaluation metrics, followed by providing implementation details. Subsequently, we compare ICVL with state-of-the-art methods under various popular benchmarks, and finally present ablation studies of the proposed strategies.

\subsection{Datasets and Evaluation Metrics}
\paragraph{Ego4D} \cite{ego4d:grauman2022ego4d}. This dataset is a large-scale egocentric dataset encompassing hundreds of scenarios, such as home, outdoor, and workplace environments. We conduct experiments on its \textit{Forecasting} subset, which includes a total of 243 hours of video, 3472 annotated clips. It has 117 verbs and 521 nouns for the LTA task. We adhere to the dataset's standard splits for evaluation.

\paragraph{EPIC-KITCHENS-55 (EK-55)} \cite{ek55:damen2020epic}. This dataset contains 55 hours of egocentric videos centered around cooking scenarios, recorded by 32 participants in 32 different kitchens. It contains 125 verb categories and 352 noun categories. We follow the splits provided by \cite{egotopo:nagarajan2020ego}.

\paragraph{EGTEA Gaze+ (EGTEA)} \cite{egtea:li2018eye}. This dataset is a first-person dataset containing 86 densely labeled cooking videos over 26 hours, with 19 verb categories and 51 noun categories. We also follow the splits provided by \cite{egotopo:nagarajan2020ego}.

\paragraph{Evaluation Metrics.} For Ego4D, we employ the default edit distance (ED) metric using the Damerau-Levenshtein distance \cite{ed:damerau1964technique}. ED is computed separately for verbs, nouns, and actions sequences. Given an observed video with \textit{N}$_{seg}$ = 8, we report the minimum edit distance between \textit{K} = 5 predicted sequences, each of length \textit{Z} = 20. For the EK-55 and EGTEA datasets, we follow the setting in \cite{egotopo:nagarajan2020ego}, using mean average precision (mAP) for multi-label classification as the evaluation metric. The task involves observing the first \(P\%\) of each video and predicting the actions that will occur in the remaining \((100-P)\%\) of the video. Here actions are defined to verbs only. We consider \(P=[25,50,75]\) to represent different anticipation horizons and report performance on the validation set for all target actions (All), frequently appeared actions (Freq), and rarely appeared action (Rare) respectively.

\subsection{Implementation Details}
For action recognition, we utilize the frozen encoder CLIP ViT-L/14 to extract visual features and then employ a Transformer encoder with 8 attention heads. For ICAF module, we utilize BLIP2-OPT-2.7B \cite{blip2:li2023blip} as the frozen visual encoder, LLaMA 3.2-9B as the VLM to derive behavioral intentions, along with LLaMA 3-8B \cite{llama3:dubey2024llama} as the text encoder and the LLM for anticipation. The Adam optimizer is used for end-to-end training with a learning rate of $5 \times 10^{-5}$, over 8 epochs.

\subsection{Results and Analysis}
\paragraph{Comparison to  state-of-the-art.} We compare ICVL with the current state-of-the-art approaches. Table \ref{tab:ego4d} shows the performance comparison on the Ego4D dataset where our method consistently outperforms the previous SOTA \cite{palm:kim2025palm,egovideo:pei2024egovideo} in terms of edit distance for noun, verb, and action with an improvement of \{2.71\%, 5.95\%, 2.49\%\} and \{0.7\%, 0.6\%, 0.49\%\}, respectively. Notably, most methods using LLMs need  to obtain the observed action labels, and the accuracy of action recognition models varies with different visual encoders. Specifically, the action recognition accuracy is 7.97\% for CLIP encoder \cite{clip:radford2021learning}, 20.63\% for EgoVLP encoder \cite{egovlp:lin2022egocentric}, and 27.64\% for EgoVideo encoder \cite{egovideo:pei2024egovideo}. A direct relationship between the recognition accuracy and the final anticipation performance can be clearly observed. Results show that ICVL achieves a significant performance improvement of \{5.61\%, 2.12\%, 3.61\%\} over other approach using the same CLIP encoder \cite{antgpt:zhao2023antgpt}. Additionally, it still outperforms methods that employ stronger visual encoders, delivering the best anticipation performance overall. This indicates that our method is more robust and reliable, where the learned intention-enhanced visual embeddings and selected examples effectively mitigate the noise of observed action labels. Among the LLM-based methods, AntGPT, PlausiVL, PALM, EgoVideo only use textual inputs while PlausiVL focuses only on the original visual embeddings. Our method emphasizes the integration of information from both modalities, demonstrating that LLMs can achieve accurate predictions by leveraging enhanced visual features and carefully designed textual prompts.

Table \ref{tab:ek} presents a comparison between our method and previous state-of-the-art approaches on the EK-55 and EGTEA datasets. Our method achieves the best performance on both datasets, with particularly notable results on EK-55 dataset, showing an improvement of 2.9\%, 2.3\% and 1.9\% on all actions, frequently happened actions and rarely happened actions respectively.

\subsection{Ablation Studies}

\paragraph{Effectiveness of the two proposed modules.} The results on the Ego4D dataset of ICAF and Example Selection modules are provided in Table \ref{tab:ablation_study}. It is evident that both modules contribute to a significant overall improvement in model performance, with ICAF having the greatest impact. This is primarily because intentions can enhance the extraction of discriminative information from visual features, providing critical visual cues for actions' evolution and aiding LLMs in making predictions. Additionally, the carefully selected examples also enrich the inputs to the LLM, enhancing the in-context learning ability of the LLM.

\begin{figure}[t]
	\centering
	\includegraphics[width=0.71\columnwidth]{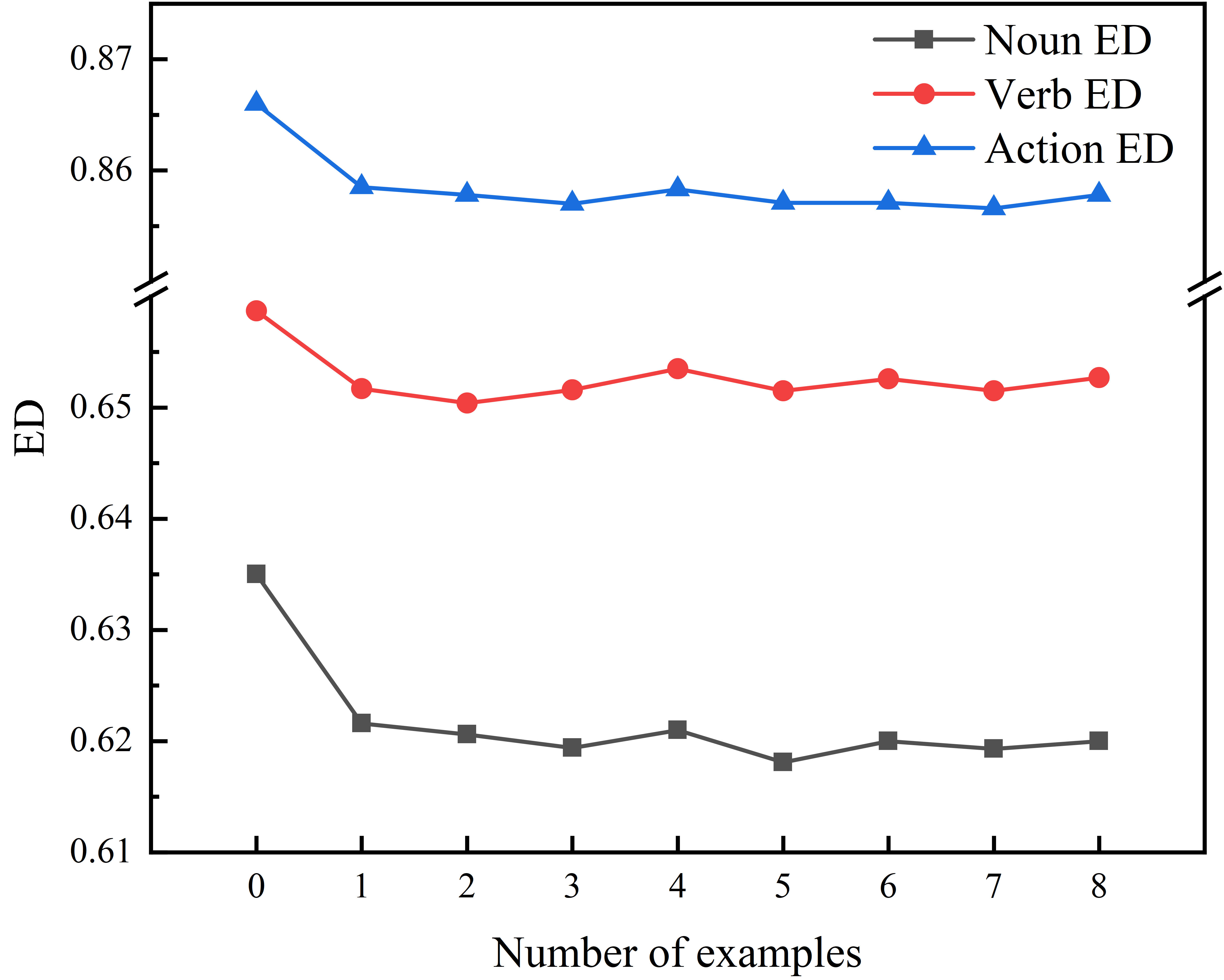} 
	\caption{Ablation study on the number of the Selected Examples.}
	\label{fig:example}
\end{figure}

\begin{table}[t]
	\centering
	\begin{tabular}{@{}lccc@{}}
		\toprule
		Method & Noun $\downarrow$ & Verb $\downarrow$ & Action $\downarrow$ \\ 
		\midrule
		Baseline & 0.6927 & 0.6823 & 0.8944 \\ 
		Baseline w/ ES & 0.6549 & 0.6759 & 0.8813 \\
		Baseline w/ ICAF & 0.6287 & 0.6550 & 0.8643 \\ 
		\hline
		ICVL & \textbf{0.6194} & \textbf{0.6516} & \textbf{0.8570} \\  
		\bottomrule
	\end{tabular}
	\caption{Ablation study on ICAF and Example Selection (ES). Baseline refers to fine-tuning LLM only with text-prompt input.}
	\label{tab:ablation_study}
\end{table}

\begin{table}[t]
	\centering
	\begin{tabular}{@{}lccc@{}}
		\toprule
		Derivation & Noun $\downarrow$ & Verb $\downarrow$ & Action $\downarrow$ \\ 
		\midrule
		Baseline & 0.6469 & 0.6661 & 0.8773 \\
		Visual features & 0.6454 & 0.6580 & 0.8733 \\
		Action labels & 0.6383 & 0.6605 & 0.8680 \\ 
		VLM & \textbf{0.6287} & \textbf{0.6550} & \textbf{0.8643} \\ 
		\bottomrule
	\end{tabular}
	\caption{Ablation study on intention generation. Baseline refers to using visual embeddings without intention integration.}
	\label{tab:intention}
\end{table}

\begin{table}[t]
	\centering
	\begin{tabular}{@{}lccc@{}}
		\toprule
		Method & Noun $\downarrow$ & Verb $\downarrow$ & Action $\downarrow$ \\ 
		\midrule
		Concat & 0.6329 & 0.6610 & 0.8676 \\
		CrossAttn (V) & \textbf{0.6285} & 0.6580 & 0.8656 \\
		CrossAttn (I) & 0.6287 & \textbf{0.6550} & \textbf{0.8643} \\ 
		\bottomrule
	\end{tabular}
	\caption{Ablation study on the integration between intention and visual embeddings.}
	\label{tab:cross}
\end{table}

\paragraph{Design of ICAF.} We examine the design of the ICAF module from two aspects: the approach to generating intentions and the integration between intention and visual embeddings. Results concerning the ways of generating intentions on the Ego4D dataset are reported in Table \ref{tab:intention}, of which \textit{Visual features} represents generating latent intention embeddings solely from visual features using learnable tokens, \textit{Action labels} means generating intention from observed action labels via LLM, and \textit{VLM} refers to generating intentions through a VLM. Compared with \textit{Baseline} without using intentions, results show that integrating intentions effectively enhances the discriminative power of visual features and improves LLM predictions. Using a VLM to infer intentions from video inputs achieves the best performance. 

Table \ref{tab:cross} demonstrates the impact of different integration strategy. \textit{Concat} refers to concatenating the intention and visual embeddings. \textit{CrossAttn (V)} denotes employing visual embeddings as the query in cross-attention method, whereas \textit{CrossAttn (I)} utilizes intention embeddings as the query. Results illustrate that the cross-attention methods are superior to the method of simple concatenation. This indicates that cross-attention method successfully integrates intention embeddings into the visual embeddings, allowing the visual cues relevant to the intentions to be highlighted, thus obtaining intention-enhanced visual embeddings. Additionally, the choice of modality for the query affects performance, with using intention as the query yielding the best results.

\paragraph{Influence of the number of examples.} Figure \ref{fig:example} shows the impact of the number of examples on the Ego4D dataset. As observed, providing a high-quality example significantly improves model performance. However, the relative performance improvement decreases as the number of examples increases. Once the number of examples reaches three, further

\begin{table}[H]
	\centering
	\begin{tabular}{@{}cccc@{}}
		\toprule
		Examples  & Noun ED $\downarrow$ & Verb ED $\downarrow$ & Action ED $\downarrow$ \\
		\midrule
		
		Text Similarity & 0.7299 & 0.6994 & 0.9076  \\
		Visual Similarity & 0.7330 & 0.6948 & 0.8993 \\ 
		Fused Similarity & \textbf{0.7128} & \textbf{0.6646} & \textbf{0.8912} \\
		\bottomrule
	\end{tabular}
	\caption{Ablation study on the example selection method.}
	\label{tab:example}
\end{table}

\noindent increases have a negligible effect on performance, with optimal results attained when seven examples are provided. This is likely due to the decreasing relevance of subsequent examples, which fail to provide more meaningful guidance to the LLM. We report the final performance based on the use of three examples.

\paragraph{Effectiveness of example selection method.} Table \ref{tab:example} illustrates the impact of example selection based on different modalities on the Ego4D dataset. The results indicate that examples selected based on the visual modality outperform those chosen based on textual similarity, likely due to the authenticity of the visual information. Our multimodal selection strategy, which considers both modalities, identifies the most relevant examples, proving the effectiveness of our approach.

\section{Limitation and Conclusion}
In this study, we explore how to effectively utilize both visual and textual modalities through LLM to tackle LTA tasks. To make visual features more discriminative, we introduce an Intention-Context Attention Fusion mechanism that integrates visual embeddings with behavior intentions inferred by VLM. Furthermore, to improve the LLM's understanding of the task and enhance its in-context learning capabilities, we propose a multi-modality example selection mechanism that provides more relevant examples. Extensive experiments on Ego4D, EPIC-Kitchens-55 and EGTEA GAZE+ datasets validate the effectiveness of our Intention-Conditioned Vision-Language model. While this work represents a preliminary investigation into multimodal fusion method using LLM, future research may focus on improving the logical consistency and coherence of action sequences predicted by the LLM.

\bibliographystyle{named}
\bibliography{ijcai25}

\end{document}